\documentclass[conference,compsoc]{IEEEtran}
\usepackage{multirow}
\usepackage{multicol}
\usepackage{float}
\usepackage{threeparttable}

\ifCLASSOPTIONcompsoc
  \usepackage[nocompress]{cite}
\else
  \usepackage{cite}
\fi

\usepackage{xcolor,colortbl}

\definecolor{LighterGreen}{rgb}{0.8, 1, 0.8}
\definecolor{LightGreen}{rgb}{0.6, 1, 0.6}
\definecolor{LighterRed}{rgb}{1, 0.9, 0.9}
\definecolor{LightRed}{rgb}{1, 0.6, 0.6}

\ifCLASSINFOpdf
  \usepackage[pdftex]{graphicx}
  \graphicspath{{./figures/}}
  \DeclareGraphicsExtensions{.pdf,.jpeg,.png}
\else
\fi

\usepackage{amsmath}

\usepackage{authblk}
\usepackage{amsfonts}
\usepackage{booktabs}
\usepackage{enumitem}

\newcommand{\equref}[1]{Eq.~\(\ref{#1}\)}
\newcommand{\figref}[1]{Figure~\ref{#1}}
\newcommand{\secref}[1]{Section~\ref{#1}}
\newcommand{\tabref}[1]{Table~\ref{#1}}

\usepackage{pifont}
\newcommand{\xmark}{\ding{55}}%

\usepackage{url}

\begin{document}

\title{When Simple Model Just Works: Is Network Traffic Classification in Crisis?\\
\thanks{Identify applicable funding agency here. If none, delete this.}
}

\author[2]{Kamil Jerabek}
\author[1,3]{Jan Luxemburk}
\author[1,3]{Richard Plny}
\author[1,3]{Josef Koumar}
\author[1,3]{Jaroslav Pesek}
\author[1,3]{Karel Hynek}
\affil[1]{Czech Technical University in Prague, Faculty of Information Technology, Czech Republic}
\affil[2]{Brno University of Technology, Faculty of Information Technology, Czech Republic}
\affil[3]{CESNET, a.l.e., Prague, Czech Republic \authorcr Corresponding author email: {\tt ijerabek@fit.vutbr.cz}}


\maketitle

\begin{abstract}
Machine learning has been applied to network traffic classification (TC) for over two decades. While early efforts used shallow models, the latter 2010s saw a shift toward complex neural networks, often reporting near-perfect accuracy. However, it was recently revealed that a simple k-NN baseline using packet sequences metadata (sizes, times, and directions) can be on par or even outperform more complex methods. In this paper, we investigate this phenomenon further and evaluate this baseline across 12 datasets and 15 TC tasks, and investigate why it performs so well. Our analysis shows that most datasets contain over 50\% redundant samples (identical packet sequences), which frequently appear in both training and test sets due to common splitting practices. This redundancy can lead to overestimated model performance and reduce the theoretical maximum accuracy when identical flows have conflicting labels. Given its distinct characteristics, we further argue that standard machine learning practices adapted from domains like NLP or computer vision may be ill-suited for TC. Finally, we propose new directions for task formulation and evaluation to address these challenges and help realign the field.
\end{abstract}


\IEEEpeerreviewmaketitle

\section{Introduction}

Traffic classification (TC) based on machine learning (ML) has become essential in network security research, particularly with the increasing prevalence of encrypted communication. As traditional methods like deep packet inspection lose effectiveness, ML provides a viable and often the only alternative for identifying communication patterns. Advances in deep representation learning in fields like computer vision and natural language processing have inspired similar approaches for traffic classification. Recent work by Luxemburk et al.~\cite{Luxemburk2025Universal} proposed a universal flow representation and unexpectedly found that a simple baseline based only on packet sequence and a k-NN classifier performed on par with or exceeded state-of-the-art (SOTA) methods across multiple datasets. We refer to this baseline as \textit{input-space}, reflecting its use of raw packet sequence metadata without any additional feature processing. Nevertheless, the mentioned study only briefly addressed the findings and emphasized the need for further research.

\begin{figure}[t!]
\centering
\includegraphics[width=\linewidth]{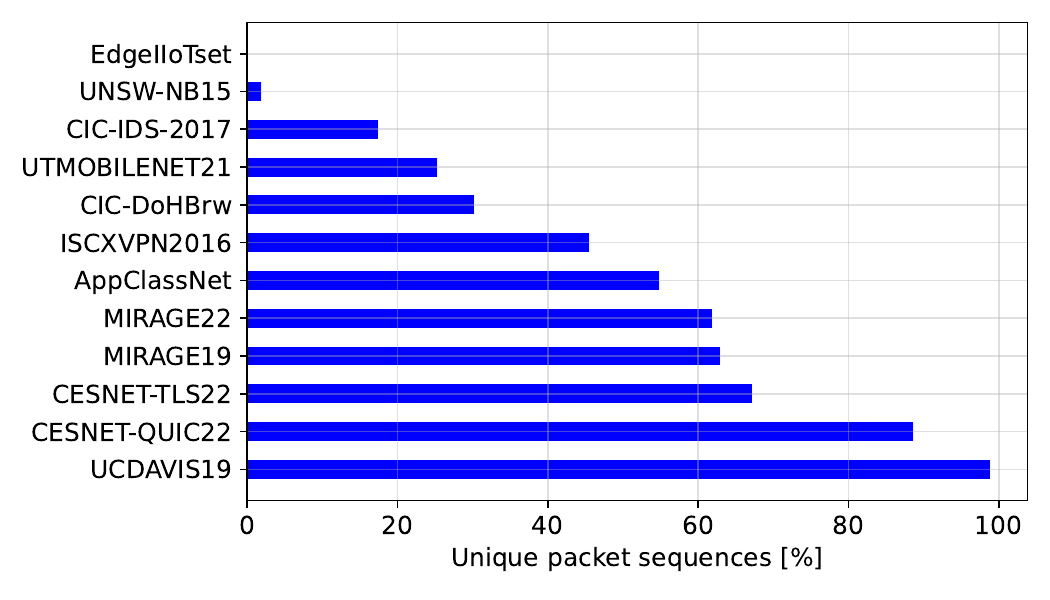}
\vspace{-0.7cm}
\caption{Fraction of unique sequences of packet lengths and directions present in the examined datasets.}
\label{fig:unique_flow_overall}
\end{figure}

In this work, we replicate and analyze the results of the simple input-space baseline across 12 datasets and 15 TC tasks, and discover that its strong performance can be largely explained by the presence of extensive redundancy within the datasets. As shown in \figref{fig:unique_flow_overall}, half of the evaluated datasets would shrink to less than 50\% of their original size if duplicates were removed and only unique samples were counted. We also show how such redundancy affects evaluation practices by estimating the theoretical maximum accuracy achievable per dataset. 

Our findings raise concerns about the relevance of current evaluation methodologies in the TC field. Unlike other domains such as computer vision or NLP, network traffic originates mostly from API requests, which may vary in content but typically share request-response similarities on the packet level. Traditional evaluation approaches, however, fail to account for this characteristic. To address this limitation, we propose recommendations for adapting evaluation protocols to produce more trustworthy and representative results.

The paper is organized as follows. Section~\ref{sec:related-works} describes advancements in the field prior to this work. Section~\ref{sec:exploring-knn-performance} defines the baseline method and presents results on 12 datasets. Section~\ref{sec:reason-exploration} provides an in-depth analysis of the network traffic data. The outcomes and recommendations are covered in the discussion Section~\ref{sec:discussion}. The last Section~\ref{sec:conclusion} concludes the work.

\section{Related Works}
\label{sec:related-works}

The network traffic classification has experienced significant evolution over time. In the early 2000s, researchers began applying machine learning techniques at a larger scale to network traffic. Several surveys~\cite{nguyen2009survey,boutaba2018comprehensive} describe this pioneering period, during which the community explored various approaches and candidate features, typically categorized into flow-level and packet-payload features. The flow-level features used in~\cite{Williams2006Preliminary,Karagiannis2005BLINC} contain information directly from the flow, such as IP addresses, number of transmitted bytes and packets, and other statistical information gathered from the packet sequence. The payload features used in~\cite{Wang2004} typically contain information from the L7 protocol headers. The rise of traffic encryption has reduced the relative importance of payload features, but they still remain valuable in some scenarios.

Despite no general consensus on the type of universal traffic classification features, the research community in~\cite{bernaille2006traffic,Dewes2003,crotti2007traffic} has gradually adopted time-series information derived from the initial packets of each flow as one of the primary types of features. These features, often referred to as the Sequence of Packet Lengths and Times (SPLT), comprise the lengths of individual packets, their direction (inbound vs. outbound), and the inter-packet arrival times. 

In the latter half of the 2010s, TC has moved to the deep learning era. Numerous studies~\cite{Yang2021,chen2016automatic,wang2017end,wang2024data,akbari2023critical,malekghaini2022data} leveraged recent advances in data science to enhance classification performance, applying deep learning techniques to a variety of network traffic scenarios. These methods also use the combination of all the features previously mentioned. Several datasets were introduced at that time, some of which have since become de facto standards in the research community. Important examples include ISCXVPN2016~\cite{dataset_ISCXVPN2016}, UNSW-NB15~\cite{dataset_UNSWNB15}, and CIC-IDS-2017~\cite{dataset_CICIDS2017}, each cited thousands of times.

However, beyond their age, these datasets are known to contain significant issues. For instance, Lanvin et al.~\cite{lanvin2022errors} identified critical flaws in the CIC-IDS-2017 dataset, including inconsistent timestamps, mislabeled samples, and duplicate flows. Although they released a corrected version, the community continues to predominantly rely on the original dataset. Other studies, aware of these problems, have performed custom filtering to mitigate the issues. Aceto et al.~\cite{aceto2021distiller} analyzed the ISCXVPN2016 dataset and identified several inconsistencies, providing a detailed description of the filtering steps required to make it usable. Nevertheless, the majority of subsequent studies continue to rely on the original, unfiltered version of the dataset.

Despite acknowledged concerns regarding dataset quality---as previously described and also discussed in several surveys~\cite{Alex2023,goldschmidt2025network}---the research community has continued to evaluate deep learning models primarily using the data in the original form. Classical shallow machine learning methods have been largely overlooked, amid a prevailing consensus that more sophisticated models are essential to achieve further performance gains. More recently, Luxemburk et al.~\cite{Luxemburk2025Universal} compared their novel embedding-based approach with a simple input-space baseline. Surprisingly, the baseline often matched or even outperformed more complex deep learning methods. However, the study did not thoroughly investigate the underlying reasons for the baseline effectiveness, nor was the evaluation extended across multiple datasets.

In this study, we adopt the baseline method proposed by Luxemburk et al.~\cite{Luxemburk2025Universal}. We evaluate its performance across 12 widely used datasets from the traffic classification domain. Furthermore, we conduct an in-depth analysis to investigate the underlying factors contributing to the unexpectedly high performance of this simple baseline. Based on our findings, we offer recommendations for refining experimental protocols within the TC research community.

\section{Exploring k-NN Performance}
\label{sec:exploring-knn-performance}

In this section, we summarize the baseline approach~\cite{Luxemburk2025Universal} and evaluate its performance over 15 classification tasks. To ensure a fair comparison, all experiments are conducted under consistent conditions using standardized evaluation metrics. Moreover, we apply Optuna-based hyperparameter tuning to optimize selected parameters and establish the baseline performance limits. Our goal is to validate the strength of this baseline as an encrypted traffic classification method and highlight some significant limitations of popular TC datasets.

\begin{table*}
  \centering
  \caption{Public datasets and evaluation splits used in our experiments.}
  \label{tab:datasets-description}
  \begin{tabular}{l l r r l r l}
    \toprule
    \textbf{Dataset} & \textbf{Task} & \textbf{\#Flows} & \textbf{\#Classes} &
    \textbf{Splits (Train / Val / Test)} & \textbf{\#Reps} & \textbf{Split strategy} \\
    \midrule
    CESNET-TLS22~\cite{Luxemburk2023TLS} & TLS  & 141M & 191 & 1M / 0.1M / 1M  & 5 & time-based (weeks 1 vs 2) \\
    CESNET-QUIC22~\cite{Luxemburk2023QUICDataset} & QUIC & 153M   & 102 & 1M / 0.1M /1 M  & 5 & time-based (weeks 1 vs. 2,3,4) \\
    ISCXVPN2016~\cite{dataset_ISCXVPN2016}        & VPN & 10.5k  & 2 & 60\% / 20\% / 20\% & 5 & random            \\
    MIRAGE19~\cite{dataset_MIRAGE19}              & Android apps & 122k & 20  & 80\% / 10\% / 10\% & 5 & prepared random splits   \\
    MIRAGE22~\cite{dataset_MIRAGE22}              & Video-meeting apps & 59k & 9   & 80\% / 10\% / 10\% & 5 & prepared random splits   \\
    UTMOBILENET21~\cite{dataset_UTMOBILENET21}    & Mobile apps                   & 9.5k   & 17  & 80\% / 10\% / 10\% & 5 & prepared random splits   \\
    UCDAVIS19~\cite{dataset_UCDAVIS19}            & QUIC                 & 7k & 5   & 7k / --- / 83 or 150 & 1 & fixed           \\
    AppClassNet~\cite{dataset_AppClassNet}        & TLS                      & 10M    & 500 & 1M / 0.1M / 1 M  & 5 & random            \\
    EdgeIIoTset~\cite{dataset_EdgeIIoTset}        & IIoT                & 73M     & 15 & 1M / 0.1M / 1M  & 5 & random            \\
    UNSW-NB15~\cite{dataset_UNSWNB15}             & Network attacks          & 2.5M   & 10 & 0.5M / 0.1M / 0.5M & 5 & random        \\
    CIC-IDS-2017~\cite{dataset_CICIDS2017}        & Network attacks & 2M & 8  & 0.5M / 0.1M / 0.5M & 5 & random        \\
    CIC-DoHBrw~\cite{dataset_CICDoHBrw2020}       & DoH                & 1.5M & 2 & 0.5M / 0.1M / 0.5M & 5 & random        \\
    \bottomrule
  \end{tabular}
\end{table*}

\subsection{Definition of Input-Space Baseline}
\label{subsec:input_space_def}
Input-space baseline is a k-NN classifier using L1 distance and the following feature set: sizes, inter-packet times in milliseconds (IPT), and directions (encoded as \textpm1) of the first \textit{N} (up to 30) payload packets. The use of the L1 metric measures distances through simple sums of differences at individual packet positions and features. Of course, this means that the baseline cannot ever capture differences across packet positions (such as missed and retransmitted packets or shifted patterns in the sequences). As the three packet features have different units and scales, we define the following four parameters to allow their reweighting:

\begin{enumerate}
    \item $N$ -- the number of packets;
    \item $DIR_{scale}$ -- the scaling factor for directions;
    \item $IPT_{scale}$ -- the scaling factor for IPTs;
    \item $IPT_{maxclip}$ -- IPTs are clipped to this maximum value before scaling.
\end{enumerate}

The specific configuration used in~\cite{Luxemburk2025Universal} was $N = 10$, $DIR_{scale} = 1$, $IPT_{scale} = \frac{1}{10}$, and $IPT_{maxclip} = 1000$. In this work, for each dataset, we use its validation set and the Optuna framework~\cite{Akiba2019Optuna} to find a baseline configuration with the best performance. We acknowledge that the parameter optimization process, which is described in more detail in~\secref{sec:optuna-tuning}, makes the baseline more complex. However, in comparison to black-box models like CNNs, it still remains simple and straightforward to interpret.

\subsection{Evaluation Protocol}
We selected 12 traffic classification or IDS datasets to evaluate the input-space baseline. Some datasets provide more tasks in the form of multiple separate test sets (UCDAVIS19) or multiple labels (ISCXVPN2016). In total, we have 15 classification tasks; for each, we use the same evaluation protocol explained below.

First, we put in extra effort to obtain the same data as that used in SOTA and follow the same filtering and preprocessing steps. We used all datasets available in the tcbench framework\footnote{https://tcbenchstack.github.io/tcbench/}, which offers curated and pre-split versions of popular datasets. For ISCXVPN2016, we contacted the authors of the current best SOTA~\cite{Nascita2023Embeddings}, who sent us their preprocessed version. For the remaining datasets, we did our best to choose the most relevant SOTA and follow its preprocessing steps. The exact train/test/split procedure of each dataset is described in~\secref{sec:datasets}.

With the data available, preprocessed, and split into train/validation/test sets, we run five repetitions and average the results. In each iteration, the training set is used to train a k-NN classifier, which we implement as a Faiss~\cite{Johnson2019Faiss} index that provides an efficient nearest neighbor search. Then, for each test sample, we obtain a prediction as the label of the closest (top1) training sample. Apart from this \textit{top1} approach, we also evaluate distance-based majority voting that is detailed in~\secref{sec:distance-majority}.

\subsection{Datasets}
\label{sec:datasets}

Twelve traffic classification datasets focused on various tasks were chosen for comparison with SOTA. The dataset selection was limited to only those that were published in a format where packet sequences of network flows were obtainable; for example, PCAP format or flow records. When the dataset was provided in PCAP format, we used ipfixprobe\footnote{\url{https://github.com/CESNET/ipfixprobe}} flow exporter to extract the packet sequences. In this section, we describe the selected datasets and provide a structured overview in~\tabref{tab:datasets-description}.

\textbf{CESNET-TLS22} and \textbf{CESNET-QUIC22} were both collected at backbone lines spanning two and four weeks, respectively. The datasets' authors, Luxemburk et al.~\cite{Luxemburk2023TLS, Luxemburk2023QUIC}, used the whole week for training and the week or weeks after for testing; we refer to this as a \textit{time-based} train-test split. In our case, we follow the described approach, using the first week of each dataset for training and the remaining weeks (in the case of TLS, one week; in the case of QUIC, three weeks) for testing. \secref{sec:time-based-split} discusses the challenges and benefits of time-aware evaluation and examines the observed performance degradation in more detail.

\textbf{ISCXVPN2016} was published in 2016 and represents a captured real traffic generated by lab members~\cite{dataset_ISCXVPN2016}. For each traffic type, the authors captured regular sessions and sessions over VPN. Historically, researchers have used various approaches to preprocess this dataset. We appreciated the approach of Nascita et al.~\cite{Nascita2023Embeddings}, who removed broadcast flows and cleaned the dataset of other sources of noise. They provided us with their preprocessed version consisting of around 10.5k samples.

Mirage series datasets \textbf{MIRAGE19}~\cite{dataset_MIRAGE19} and \textbf{MIRAGE22}~\cite{dataset_MIRAGE22} focus on services and capture real users; the first is based on interactions with 20 Android applications, the latter is focused on video meeting applications such as Zoom, Webex, or Teams (9 applications in total). We used the prepared train/validation/test splits available in tcbench, with filtered flows shorter than 10 packets. \textbf{UTMOBILENET21} captures Android application communication, providing packet sequences in CSV format~\cite{dataset_UTMOBILENET21}. The authors of tcbench cleaned the data, assembled flows, filtered flows with less than 10 packets, and divided the remaining 9.5k samples into five splits, which we use in this work.

The first QUIC traffic dataset \textbf{UCDAVIS19} was published in 2019~\cite{dataset_UCDAVIS19} and contains five Google services: Google Drive, Google Docs, Google Search, Google Music, and  YouTube. It includes a pretraining partition and two test sets---human (83 samples) and script (150 samples). For this dataset, we decided not to use the prepared splits in tcbench. Rather, we used the entire pretraining as the training set and evaluated the two test sets (which we consider as separate classification tasks) without averaging over multiple repetitions.

\textbf{AppClassNet} from 2022 contains TLS flows drawn from 500 mobile and web applications. Authors of the dataset, Wang et al.~\cite{dataset_AppClassNet} provide an official split, but for consistency we averaged over five repetitions.

Industrial IoT (IIoT) dataset \textbf{EdgeIIoTset}~\cite{dataset_EdgeIIoTset} mixes legitimate IIoT telemetry (MQTT) with diverse network attacks in the same capture within a lab setup. Normal operation and attack campaigns were run intermittently between 21 November 2021 and 10 January 2022. IDS dataset \textbf{UNSW-NB15} was captured in 2015 at UNSW Canberra’s Cyber Range Lab~\cite{dataset_UNSWNB15}. Benign traffic was replayed through a live test network while nine attack types were injected with a generator. The authors exported the training and test sets; however, for consistency, we generated our own splits. The very popular IDS dataset \textbf{CIC-IDS-2017}~\cite{dataset_CICIDS2017} captures five days of generated traffic with seven types of distinct attacks. Background benign traffic is generated by 25 synthetic employees via tool B-Profile, and the malicious traffic is generated by 15 individual tools. The authors used a star topology with the victim in the center. We filtered out empty flows and broadcasts. The DoH dataset \textbf{CIC-DoHBrw-2020} contains encrypted flows labeled as non-DoH, benign-DoH, or malicious-DoH, created by MontazeriShatoori et al.~\cite{dataset_CICDoHBrw2020}. It captures automated browsing of the Alexa Top-10k in Google Chrome and Mozilla Firefox. Malicious DoH traffic is generated in parallel by three distinct tools. The authors place a dedicated DoH proxy in the center of a star topology.

\subsection{Optuna Tuning}
\label{sec:optuna-tuning}
For all datasets---except for UCDAVIS19, which does not have a validation set---we search for the best input-space baseline configuration. We use Optuna, which implements state-of-the-art hyperparameter search algorithms, to optimize validation classification performance averaged over five train/validation/test splits. The best-found parameters are then used for evaluation on the test sets. The results of the optimized baseline are presented in~\tabref{tab:main-results}. In the \textit{Optim.} column, we can observe that the gains from optimization are modest---ranging from 0.5 to 1 percentage point for traffic classification datasets---and negligible for IDS datasets. AppClassNet is an outlier, with optimization yielding an improvement of nearly 5 percentage points over the “default” input-baseline parameters reused from~\cite{Luxemburk2025Universal}. The best configurations for each dataset are presented in~\tabref{tab:optuna-parameters}. 

\begin{table}[b]
    \centering
    \caption{The optimized baseline parameters.}
    \label{tab:optuna-parameters}
    \begin{tabular}{l l l l l }
    \toprule
        \textbf{Dataset} & $N$ & $DIR_{scale}$ & $IPT_{maxclip}$ & $IPT_{scale}$ \\
        \midrule
        CESNET-TLS22 & 7 & 80 & 2250 & 0.045 \\
        CESNET-QUIC22 & 17 & 102.5 & 4400 & 0.015 \\
        ISCXVPN2016-A. & 5 & 115 & 1650 & 0.015 \\
        ISCXVPN2016-T. & 7 & 135 & 850 & 0.555 \\
        ISCXVPN2016-E. & 6 & 5 & 750 & 0.64 \\
        MIRAGE19 & 9 & 90 & 200 & 0.03 \\
        MIRAGE22 & 8 & 42.5 & 1400 & 0.09 \\
        UTMOBILENET21 & 10 & 10 & 200 & 0.02 \\
        AppClassNet & 20 & 137.5 & - & - \\
        EdgeIIoTset & 14 & 132.5 & 550 & 0.005 \\
        UNSW-NB15 & 27 & 297.5 & 3150 & 0.44 \\
        CIC-IDS-2017 & 9 & 187.5 & 850 & 0.335 \\
        CIC-DoHBrw & 6 & 67.5 & 1400 & 0.285 \\
        \bottomrule
    \end{tabular}
\end{table}

For each dataset, a distinct configuration of hyperparameters (described in Section~\ref{subsec:input_space_def}) was selected based on validation performance. During the optimization process, the packet size sequence remained unchanged, while the other two sequence components---packet directions and inter-packet times---were scaled using multiplicative factors $DIR_{scale}$ and $IPT_{scale}$, as listed in Table~\ref{tab:optuna-parameters}. This scaling adjusted the relative influence of these features during the computation of L1 distances. Moreover, inter-packet times were clipped to a maximum value of $IPT_{maxclip}$ to reduce the impact of outliers in this feature.

The optimal relative importance of packet directions and IPTs varied across datasets, with no clear pattern. The optimal number of packets used for classification, $N$, was ten or fewer for most datasets, suggesting that the initial part of each flow carries sufficient discriminative information for identifying a wide range of traffic classes. One notable observation is that CESNET-TLS22 and CESNET-QUIC22, although being similar datasets, differ in the presence of application-level control packets (QUIC ACKs), which leads to higher optimal $N$ for CESNET-QUIC22.

\begin{table*}[ht]
    \centering
    \caption{Comparison of SOTA and input-space baseline performance across multiple datasets. All values represent an accuracy in percentage or F1 score, denoted by $\mathbb{A}$ and $\mathbb{F}_1$ respectively. Input Space -- the kNN performance with default scaling and N=10. Optim. -- the performance achieved after Optuna tuning. Distance Maj. -- the final result after additional distance-based majority voting.}
    \label{tab:main-results}
    \begin{tabular}{llllllcll}
        \toprule
        \textbf{Dataset} & \textbf{Category} & \textbf{SOTA} & \textbf{Input Space} & ~\,\,\,$\Delta$ & \textbf{Optim.} & $\Delta$ & \textbf{Distance Maj.} & \textbf{SOTA}~$\Delta$ \\
        \midrule
        CESNET-TLS22 & Web service & \cite{fauvel2023lightweight}: 97.2~~\,($\mathbb{A}$) & 90.96 & ~\,\,\,1.06 & 92.02 & 0.32 & 92.34 & \cellcolor{LightRed}$-$4.86 \\
        CESNET-QUIC22 & Web service & \cite{Luxemburk2023QUIC}: 80.87 ($\mathbb{A}$) & 55.22 & ~17.37 & 72.59 & 0.32 & 72.91& \cellcolor{LightRed}$-$7.96 \\
        ISCXVPN2016-A. & General traffic & \cite{Nascita2023Embeddings}: 79.92 ($\mathbb{A}$) & 70.94 & ~\,\,\,0.9 & 71.84 & 2.35 & 74.19 & \cellcolor{LightRed}$-$5.73 \\
        ISCXVPN2016-T. & General traffic & \cite{Nascita2023Embeddings}: 81.71 ($\mathbb{A}$) & 72.92 & ~\,\,\,0.45 & 73.37 & 2.34 & 75.71 & \cellcolor{LightRed}$-$6 \\
        ISCXVPN2016-E. & General traffic & \cite{Nascita2023Embeddings}: 93.01 ($\mathbb{A}$) & 90.61 & ~\,\,\,0.49 & 91.1 & 0.48 & 91.58 & \cellcolor{LighterRed}$-$1.43 \\
        MIRAGE19 & Mobile app & \cite{wang2024data}: 80.06 ($\mathbb{F}_1$) & 79.98 & ~\,\,\,0.68 & 80.66 & 0.12 & 80.78 & \cellcolor{LightGreen}~\,\,\,0.72 \\
        MIRAGE22 & Mobile app & \cite{wang2024data}: 97.18 ($\mathbb{F}_1$) & 95.63 & ~\,\,\,0.64 & 96.27 & 0.04 & 96.31 & \cellcolor{LighterRed}$-$0.87 \\
        UTMOBILENET21 & Mobile app & \cite{finamore2023replication}: 81.91 ($\mathbb{F}_1$) & 83.8 & $-$0.02 & 83.78 & 0.34 & 84.12 & \cellcolor{LightGreen}~\,\,\,2.21 \\
        UCDAVIS19-Script & Google service & \cite{finamore2023replication}: 98.63 ($\mathbb{A}$) & 98 &  & \xmark & & \xmark & \cellcolor{LighterRed}$-$0.63 \\
        UCDAVIS19-Human & Google service & \cite{finamore2023replication}: 80.45 ($\mathbb{A}$) & 71.08 &  & \xmark &  & \xmark & \cellcolor{LightRed}$-$9.37 \\
        AppClassNet & Web service & \cite{dataset_AppClassNet}: 88.3~~\,($\mathbb{A}$) & 76.25 & ~\,\,\,4.76 & 81.01 & 0 & 81.01 & \cellcolor{LightRed}$-$7.29 \\
        EdgeIIoTset & IDS & \cite{koumar2023network}: 99.97 ($\mathbb{A}$) & 99.79 & ~\,\,\,0 & 99.79 & 0.03 & 99.82 & \cellcolor{LighterRed}$-$0.15 \\
        UNSW-NB15 & IDS & \cite{Koumar2023NETISA}: 98.85 ($\mathbb{A}$) & 97.95 & ~\,\,\,0.15 & 98.1 & 0.21 & 98.31 & \cellcolor{LighterRed}$-$0.54 \\
        CIC-IDS-2017 & IDS & \cite{koumar2023network}: 99.93 ($\mathbb{A}$) & 99.8 & ~\,\,\,0.02 & 99.82 & 0.02 & 99.84 & \cellcolor{LighterRed}$-$0.09 \\
        CIC-DoHBrw & DoH & \cite{mitsuhashi2021identifying}: 99.81 ($\mathbb{A}$) & 98.54 & ~\,\,\,0.02 & 98.56 & 0.1 & 98.66 & \cellcolor{LighterRed}$-$1.15 \\
        \bottomrule
    \end{tabular}
    \label{tab:comparison}
\end{table*}

\begin{table*}[ht]
    \centering
    \caption{ISCXVPN2016 comparison with SOTA not using payload data as model input.}
    \begin{tabular}{l c c c }
        \toprule
        \textbf{Dataset} & \textbf{SOTA w/o Payload} & \textbf{Best Input Space} & \textbf{SOTA}$\Delta$ \\
        \midrule
        ISCXVPN2016-App & 63.92 ($\mathbb{A}$) & 74.19 & \cellcolor{LightGreen}10.27 \\
        ISCXVPN2016-TrafficType  & 65.56 ($\mathbb{A}$) & 75.71 & \cellcolor{LightGreen}10.15 \\
        ISCXVPN2016-Encapsulation & 85.45 ($\mathbb{A}$) & 91.58 & \cellcolor{LightGreen}6.13 \\
        \bottomrule
    \end{tabular}
    \label{tab:comparison2}
\end{table*}

\subsection{Distance-Based Majority Voting}
\label{sec:distance-majority}
In addition to the \textit{top1} approach, we experimented with various forms of majority voting among the nearest samples. While standard majority voting among the $k$ nearest neighbors did not yield substantial gains, distance-based voting proved more promising. Instead of voting among a fixed number of neighbors, we consider all samples within a predefined distance threshold $T_{maj}$. For each dataset, this threshold is selected based on validation performance, using a procedure similar to the one used for optimizing the input-baseline parameters. The performance gains provided by distance-based majority voting are summarized in~\tabref{tab:main-results}, with the most substantial improvements observed for the three tasks of ISCXVPN2016.

\subsection{Results}
The performance of the input-space baseline was evaluated against the highest-performing SOTA methods identified in existing literature (more details in Appendix B). To ensure a fair and direct comparison, we did our best to replicate the data processing and splitting methodology used by each selected SOTA approach.

As presented in Table~\ref{tab:comparison}, the optimized input-space baseline demonstrates competitive accuracy across a range of datasets. Accuracy was used as the primary evaluation metric, consistent with most related work, with the exception of three comparisons that report F1 scores. The results highlight that the baseline outperforms SOTA methods on two datasets, MIRAGE19 and UTMOBILENET21, and it is only 0.09\% behind on the widely used CIC-IDS-2017. Additionally, on most of the remaining datasets, the baseline performs comparably to more sophisticated models.

On the ISCXVPN2016 dataset, the baseline shows lower performance relative to the best SOTA result, which incorporates packet-payload features. However, when compared to a SOTA method that uses only packet sequences as input, the baseline significantly outperforms it, as shown in Table~\ref{tab:comparison2}. The more complex CESNET datasets, collected from a real network environment, present a greater challenge. These datasets involve time-separated captures and more diverse traffic conditions. In this context, the baseline is surpassed by SOTA models, which benefit from stronger generalization capabilities. This result underscores the value of more advanced methods in handling temporal variability and real-world heterogeneity. The UCDAVIS19 dataset also poses difficulties. While the baseline performs comparably to SOTA methods on the Script test subset---whose behavior is more closely aligned with the training set---it underperforms on the Human test subset, where behavioral variation is more pronounced. The AppClassNet dataset is a special case. It features randomized metadata from 20-packet segments across different parts of flows, making it structurally distinct. Despite this specificity, the baseline still achieves a reasonable level of accuracy.

Overall, the results indicate that this simple baseline approach can attain near-optimal performance on several datasets, and in some cases, even surpass more sophisticated SOTA methods. This finding is not an anomaly: it is consistent across various datasets and classification tasks in the field. Furthermore, it suggests that certain tasks, as currently defined by dataset creators, may not be as complex as assumed. The underlying causes of this phenomenon are examined in the following sections.

\section{Reason Exploration}
\label{sec:reason-exploration}

On average, the performance gap between input-space baseline and SOTA is $-$2.88\%. This surprisingly strong performance raises an intriguing question: \textit{How is it possible that such a simple model can get this close to complex SOTA methods?} In this section, we investigate the underlying causes and explain phenomena that have long gone unaccounted for, which in turn reveal fundamental properties of the traffic classification problems that set them apart from other machine learning domains.

\subsection{Redundant Packet Sequences}

\begin{figure}[t!]
\centering
\includegraphics[width=0.99\linewidth]{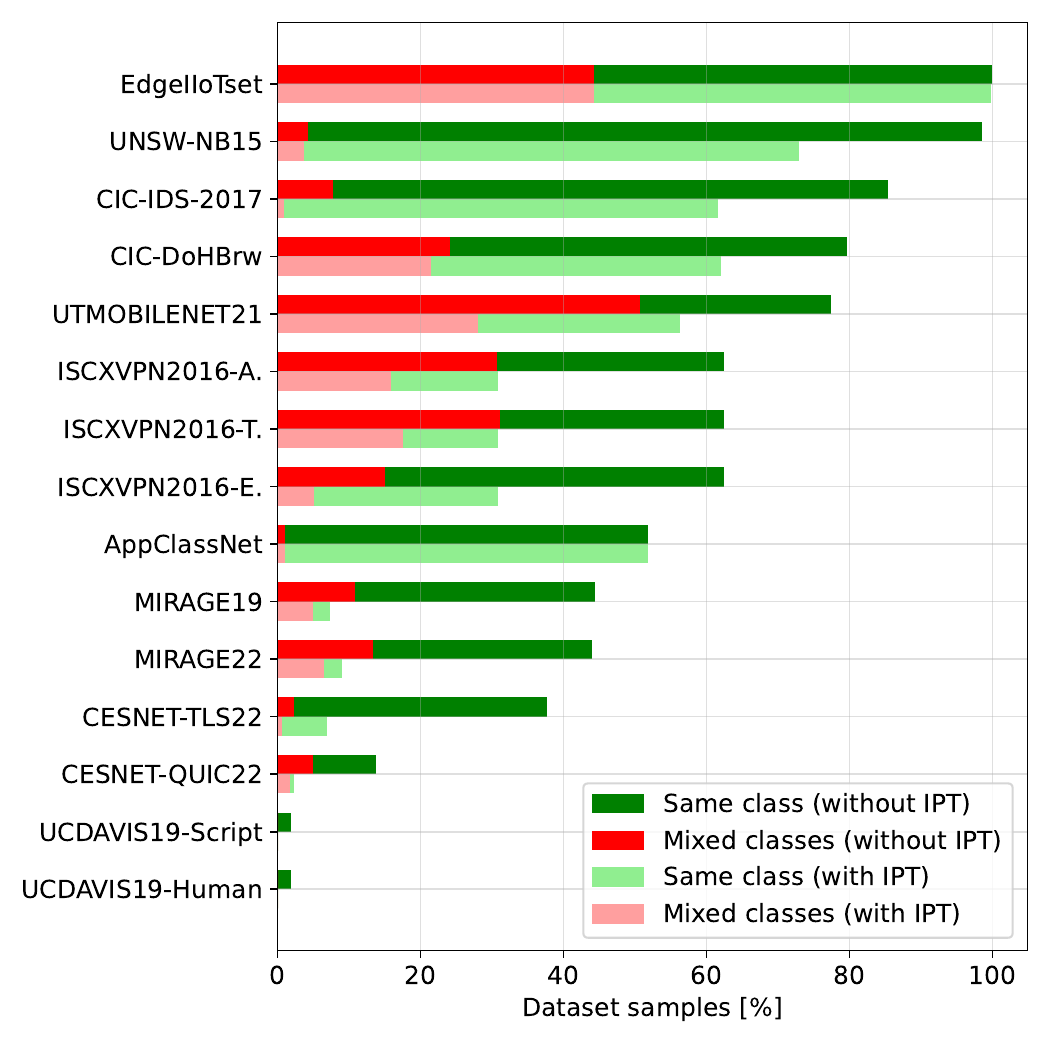}
\vspace{-0.7cm}
\caption{Fractions of duplicate samples in all datasets. Green bars represent duplicates within a single class, while red bars represent clusters of duplicates with conflicting labels. We show two variants: with and without inter-packet times in the feature vector. As expected, there are fewer duplicates when the timing information is used; however, the number of duplicates still exceeded our expectations, given that inter-packet times depend on the fluctuating network conditions and their millisecond resolution.}
\label{fig:redundant-flow-overall}
\end{figure}

We studied sample distributions and uncovered a consistent pattern across all datasets. A substantial fraction of samples are redundant as their feature vectors are exactly identical. In other words, \textbf{the datasets are full of duplicates}. The exact fractions of duplicates are shown in~\figref{fig:redundant-flow-overall}, with the overall picture being that over 50\% of samples are redundant in more than half of the datasets. Moreover, duplicate samples are not confined to a single class. Identical packet sequences appear across multiple classes, creating an unsolvable problem in which the same feature vectors are assigned conflicting labels. The implications for traffic classification can be summarized as follows:

\begin{description}
    \item[Duplicates within the same class] can make a classification task trivial, especially when datasets are split randomly into training, validation, and test sets. Identical samples end up in both training and test sets, which makes the problem trivial to solve, particularly for distance-based classifiers such as the input-space baseline defined in~\secref{subsec:input_space_def}. Possible solutions that are rarely used in related works include \textit{(1)} incorporating duplicate filtering in the evaluation pipeline of traffic classifiers, or, when proposing a new method, \textit{(2)} acknowledging the presence of duplicates and comparing performance to a suitable distance-based baseline, or \textit{(3)} avoiding random splits altogether and instead using time-based or disjoint splits, as discussed in~\secref{sec:evaluation-protocols-tc}. We believe that ignoring the high number of duplicates will lead to overconfident performance estimates and can stall the progress of traffic classification methods.

    \item[Mixed duplicates across multiple classes] pose another challenge: for a subset of such samples, correct classification is fundamentally impossible. Potential solutions include adopting multi-label approaches, relabeling duplicate samples, or adding new features to make them distinguishable. We elaborate on this issue in~\secref{sec:theoretical-max-acc}, where we also introduce a new metric for quantifying the extent of the mixed-duplicate problem: \textit{maximum achievable accuracy}. For some datasets, this upper bound can be as low as 93\%. Researchers may attempt to surpass these thresholds, unaware that doing so is inherently impossible. In other cases, we observe only a tiny gap between the \textit{maximum achievable accuracy} and the input-space baseline accuracy, indicating that the given task is effectively solved and does not require more complex methods.
\end{description}

\begin{figure*}[t]
\centering
\includegraphics[width=\linewidth]{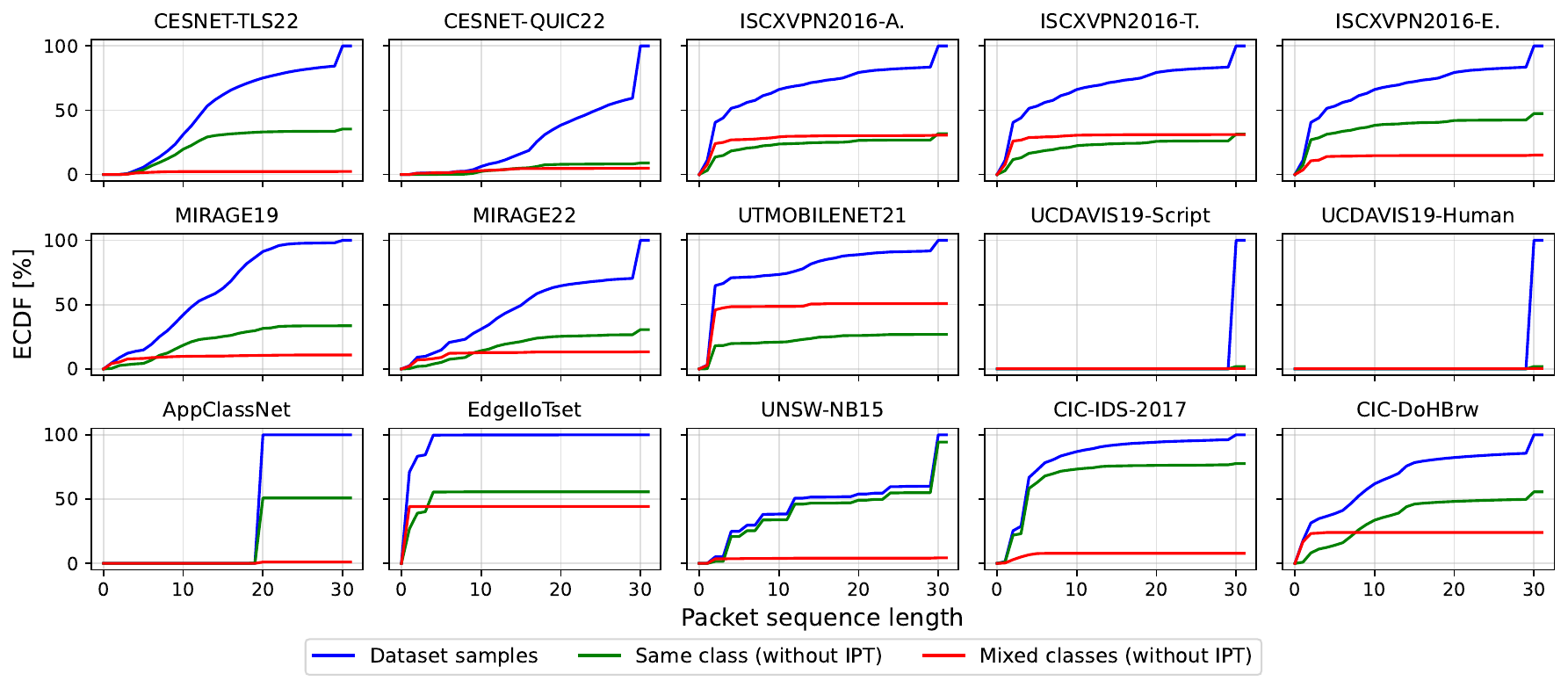}
\vspace{-0.8cm}
\caption{ECDF curves show the fractions of dataset samples, same-class duplications, and mixed-class duplications across different packet sequence lengths. The distribution of packet sequence lengths \textit{(blue)} reveals that some datasets contain fixed-size samples (AppClassNet 20 packets, UCDAVIS19 30 packets)---while others, like EdgeIIoTset and UTMOBILENET21, consist mostly of very short flows. Most mixed-class duplications \textit{(red)} originate from short sequences, whereas same-class duplications \textit{(green)} tend to mirror the overall sample distribution and are spread across all sequence lengths.}
\label{fig:ecdf_packet_sequence}
\end{figure*}

\begin{figure*}[ht!]
\centering
\includegraphics[width=0.9\linewidth]{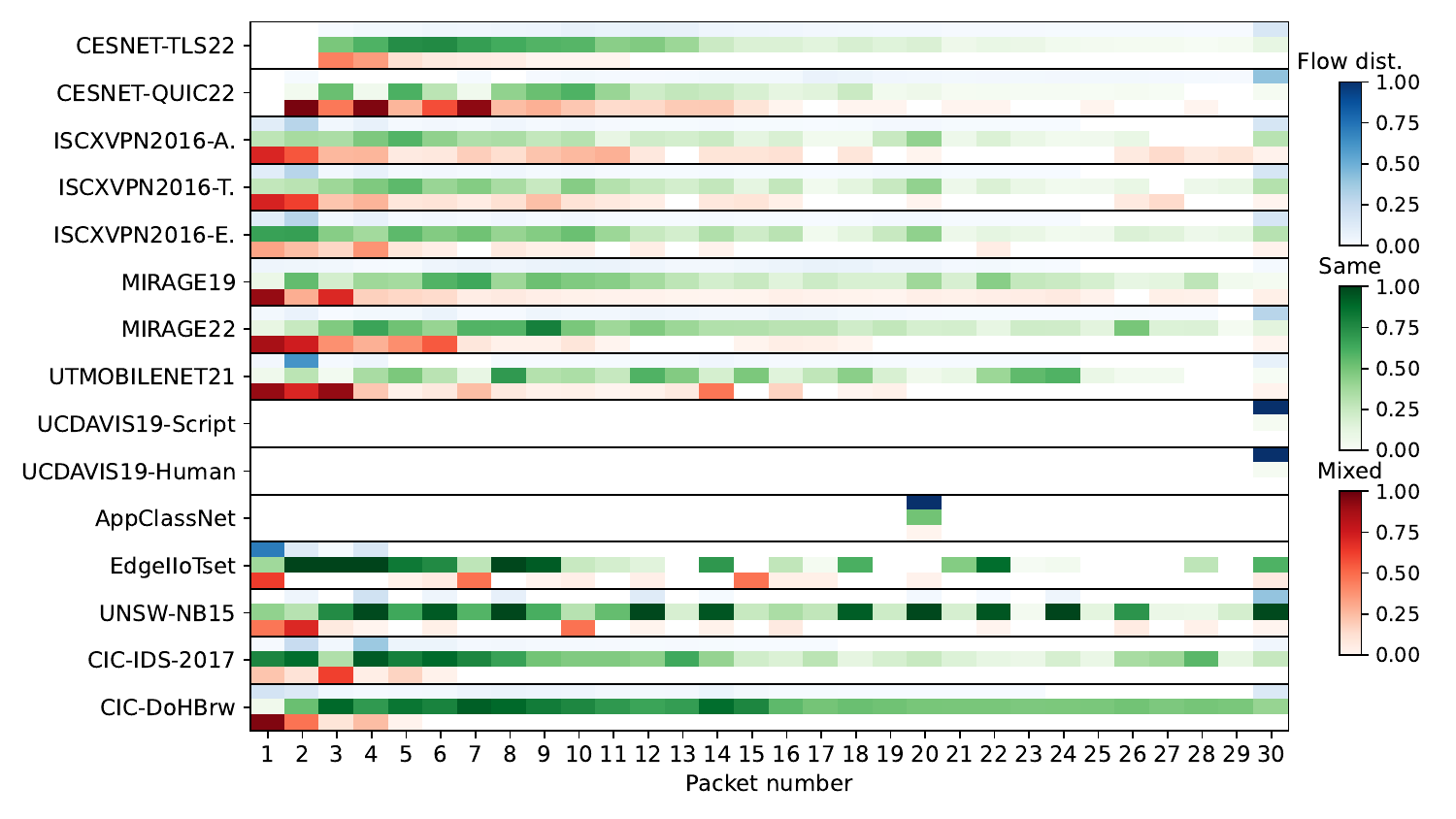}
\caption{All datasets redundancy heatmap without IPT. The color-coded redundancy ratios are calculated across all packet sequences of equal length. Blue represents a fraction of all flows in the dataset, while green (same-class duplicates) and red (mixed-class duplicates) represent the fraction of redundant samples within flows of a given length.}
\label{fig:datasets-redundancy-detail-heatmap}
\end{figure*}

A clear correlation emerges between the proportion of redundant flows (green bars in~\figref{fig:redundant-flow-overall}) and the best performance achieved with input-space baseline (the \textit{Distance Maj.} column of~\tabref{tab:main-results}). The greater the fraction of duplicates in a given dataset, the better the performance of the input-space baseline. This relationship is further evaluated using the Spearman correlation test. We set the null hypothesis as: ``There is no correlation between the performance of input-space baseline and the proportion of duplicate flow within the same class'' and chose a significance threshold of $\alpha = 1\%$. Test results are a p-value of 0.0016 and a correlation coefficient of 0.74, so we can reject the null hypothesis. We are sure that behind this confirmed correlation is the effect described earlier: duplicate samples end up in both training and test sets, which makes the classification trivial. \figref{fig:ecdf_packet_sequence} and Appendix \figref{fig:datasets-redundancy-detail-heatmap} offer a detailed view into the distribution of duplicate samples based on their packet sequence length, and the following section discusses the underlying causes of duplicate samples in network traffic.

\subsubsection{Nature of Network Traffic}
\label{sec:deterministric-communication-structure}

Encrypted network communication---when observed through the feature vector of packet metadata sequences---exhibits a rather deterministic nature that can lead to a lot of duplicate samples. Both client and server follow a predefined protocol (TLS handshake, HTTP2, QUIC); different people fetch the same resources with identical API requests; and then there are scripted bots, automatic updates, etc. All these factors contributing to duplicate samples are further amplified for short communications with a small number of exchanged packets.

On the other hand, fluctuating network conditions introduce noise, especially into inter-packet times, but can also affect packet sizes due to packet loss and retransmissions. This is often due to the client's position within the network, with factors such as physical distance, throughput, and network congestion playing a significant part. This phenomenon is evident in datasets captured from real-world environments, such as CESNET or AppClassNet datasets. In AppClassNet, Yang et al.~\cite{Yang2021} discovered that a single application can exhibit two different ``behavior profiles'' based on the residential vs. enterprise environments. The question is whether these variations are beneficial for separating traffic classes and, if not, how to train classification models to ignore them. A popular technique is data augmentations, which alter training data to produce more robust models. For instance, Xie et al.~\cite{Xie2024Rosetta} proposed a set of TCP-aware augmentations that alternate packet sequences to simulate packet loss, TCP fast retransmissions, or different MSS settings.

\begin{table}[hb!]
    \centering
    \begin{threeparttable}
    \caption{The gap between the maximum accuracy and the best performance of input-space baseline of each dataset. All metrics are averages over five folds.}
    \label{tab:max-theoretical-acc}
    \begin{tabular}{l l l l}
        \toprule
        \textbf{Dataset} & \textbf{Best Input Space} & \textbf{Max Acc.} & \textbf{Gap $\Delta$} \\
        \midrule
        CESNET-TLS22 & 92.34 & 99.41 & 7.07 \\
        CESNET-QUIC22 & 72.91 & 99.57 &  26.66 \\
        ISCXVPN2016-A. & 74.19 & 98.87 & 24.68 \\
        ISCXVPN2016-T. & 75.71 & 94.5 & 18.79 \\
        ISCXVPN2016-E. & 91.58 & 98.87 & 7.29 \\
        MIRAGE19 & 78.56\tnote{$\dag$} & 96.36 & 17.8 \\ 
        MIRAGE22 & 90.23\tnote{$\dag$} & 96.90 & 6.67 \\ 
        UTMOBILENET21 & 78.24\tnote{$\dag$} & 93.46 & 15.22 \\ 
        UCDAVIS19-Script & 98 & 100.0 & 2 \\
        UCDAVIS19-Human & 71.08 & 100.0 & 28.92 \\
        AppClassNet & 81.01 & 99.93 & 18.92  \\
        EdgeIIoTset & 99.82 & 99.85 & 0.03 \\
        UNSW-NB15 & 98.31 & 99.17 & 0.86 \\
        CIC-IDS-2017 & 99.84 & 99.97 & 0.13 \\
        CIC-DoHBrw & 98.66 & 98.79 & 0.13 \\
        \bottomrule
    \end{tabular}
    \begin{tablenotes}
    \footnotesize
    \item[$\dag$] Column \textit{Best Input Space} reuses best results from~\tabref{tab:main-results}, with the exception of MIRAGE19, MIRAGE22, and UTMOBILENET21 datasets. For these, we recomputed accuracy instead of F1-score, and did not use filtering of short flows with fewer than 10 packets.
    \end{tablenotes}
    \end{threeparttable}
\end{table}

\subsection{Mixed Duplications and Maximum Achievable Accuracy}
\label{sec:theoretical-max-acc}

We have already established that duplications mixed across multiple classes practically create a multi-label problem. Yet, hardly anyone in the traffic classification domain has ever acknowledged this or taken any countermeasures. In this section, we define a dataset metric called \textit{maximum achievable accuracy}, which takes into account samples that cannot be correctly classified. The expression in~\equref{eq:max-acc} captures the best-case scenario: it assumes a perfect model that correctly classifies all unique ($N_{unique}$), all redundant samples associated with a single class ($N_{same}$), and, for each ``cluster'' of duplicates with conflicting labels ($C_i$), it predicts the most frequent class. As defined, this metric provides an upper bound on the maximum achievable classification accuracy given the structure of the dataset.

\begin{equation}
   \label{eq:max-acc}
   \text{Max Acc.} = \frac{N_{unique} + N_{same} + \sum_{i=1}^{j} \#Maj(C_{i})}{N}
\end{equation}

We computed the maximum achievable accuracy for test sets of all datasets and compared it to the best performance of our input-space baseline. The results are presented in~\tabref{tab:max-theoretical-acc}. For most datasets, the maximum achievable accuracy falls within the range of 98–99\%; however, three popular traffic classification datasets---MIRAGE19, MIRAGE22, and UTMOBILENET21---stand out with values of 96.36\%, 96.9\%, and 93.46\%, respectively.

The gap between a simple baseline and the theoretical perfect performance represents room for improvement---a ``research playground'' for exploring more complex methods. For three particular datasets---all of them focused on IDS---CIC-DoHBrw, CIC-IDS-2017, EdgeIIoTSet, this gap is smaller than 0.15\%. This indicates that the classification tasks provided by these datasets are already solved almost perfectly by the baseline method and are therefore unsuitable for benchmarking future research proposals.

IDS tasks are known to be sensitive to the number of false positives due to the risk of alert fatigue. This makes them particularly vulnerable to the mixed-duplicates problem: duplicate samples among both benign and attack classes will inevitably cause a high number of false positives. To evaluate this effect, we estimate\footnote{For the computation of the minimal FPR, we used a slightly different approach: for each mixed cluster of benign and malware samples, we assigned the predicted label as malware, rather than using the majority class as in the computation of the maximum achievable accuracy. We believe this better reflects the behavior expected of an IDS model---namely, prioritizing the detection of malware.} the lower bound of the false positive rate (FPR) of two IDS datasets. \tabref{tab:ids-dataset-fpr} shows the estimated minimal FPR, the fraction of malicious samples in each dataset, and the fraction of malicious samples that are mixed with benign traffic. A surprising observation is that in both datasets, more than half of the malware samples are mixed up (i.e., identical samples) with benign traffic, indicating a potential issue in the annotation process. Given this degree of class overlap and the resulting high estimated FPRs, it is difficult to consider these datasets suitable for tuning a realistic intrusion detection system.

\begin{table}[t]
    \centering
        \begin{threeparttable}
        \caption{Analysis of redundancies in datasets focusing on the detection of security threats. }
        \label{tab:ids-dataset-fpr}
    \begin{tabular}{p{3.7cm}rrr}
        \toprule
        \textbf{} &  \textbf{UNSW-NB15} & \textbf{CIC-IDS-2017} \\
        \midrule
        Malicious fraction & 3.24 \% & 20.89 \% \\
         \parbox{2.15cm}{Mixed with benign}  &  60.30 \% & 52.40 \% \\  
         Minimal FPR &  2.65 \% & 9.12 \% \\ 
        \bottomrule
    \end{tabular}
    \begin{tablenotes}
    \footnotesize
    \item[$*$] We excluded the EdgeIIoTset dataset since 90\% of its samples are labeled as malicious. Due to this unrealistic fraction of malicious traffic (real environments typically have ratio \texttt{<<} 1\%), we found it unsuitable for this FPR analysis.  
    \end{tablenotes}
    \end{threeparttable}
\end{table}

\section{Discussion}
\label{sec:discussion}

A straightforward application of the k-NN algorithm on raw flow packet sequence metadata---an established data source in network traffic classification---can yield surprisingly high accuracy, closely matching sophisticated SOTA methods. As demonstrated by the results, this is not an isolated case linked to a single task or dataset. Rather, consistently high accuracy is observed across all commonly used datasets in the field.

A closer examination reveals that these datasets contain a significant degree of redundancy. In most datasets, more than $50\%$ of samples have at least one exact counterpart with an identical sequence of packet metadata values. This redundancy appears to be a characteristic of computer network communication. Importantly, this redundancy is not limited to short flows but is also observed in longer communication sequences.

Moreover, exactly the same and thus duplicate samples can also have different class labels. Such phenomena occur in all examined datasets, with the exception of the UCDAVIS19 Human and Script testing subset, which contains fewer samples and follows a different creation process. This cross-class redundancy limits the maximum achievable accuracy. Although $100\%$ accuracy is often viewed as the theoretical upper bound for machine learning models, due to such duplicates with different class labels, it is not achievable for many of the datasets. The true upper limit is often lower, leaving the researchers unknowingly chasing higher performance scores despite the inherent data limitations. 

The presented simple baseline method and the estimated upper bounds on accuracy suggest that many of the tasks defined by existing datasets offer only minimal room for improvement. This observation leads us to a central question, reflected in the title of this paper: \textit{Is the network traffic classification in crisis?} Over the past decade, numerous SOTA methods have been introduced using these datasets, yet just a few studies show meaningful improvement beyond what is achievable with a basic baseline.

The evidence presented in the paper highlights redundancy as an intrinsic characteristic of the network traffic, causing the apparent simplicity of the tasks, which are solvable by simple algorithms. However, we argue that the core issue lies in the formulation of the challenges and the design of the experimental protocol. TC research has frequently adopted experimental methodologies from other machine learning domains, such as computer vision and natural language processing. Yet, as demonstrated in this work, the TC domain---along with its datasets---constitutes a distinct branch of data science, where conventional best practices often do not work.

The typical example of the adopted evaluation protocol is the random splitting and shuffling of data samples between training and evaluation dataset parts. While such an approach is generally considered good practice for evaluating generalization, it should not be used in the TC domain. As this study highlighted, the random splitting method is not ideal in the context of network traffic data due to the high likelihood of identical or near-identical samples appearing in both training and evaluation sets. Such practice thus undermines the validity of the results, as models may effectively be tested on data they have already seen and are thus solvable by simpler algorithms. 

Based on the findings, the TC research community should thus adopt 1) more sophisticated evaluation protocols and pursue 2) novel approaches and features that would minimize the collision of the data samples from different classes.

\subsection{Evaluation Protocols of TC}
\label{sec:evaluation-protocols-tc}
Experimental protocols must account for sample redundancy as an inherent characteristic of the data and should be designed to reflect the realism of actual deployment scenarios. We can argue that duplicates between training and testing sets are not problematic when such duplicates would naturally occur in real-world usage and are not artifacts of the evaluation setup. Nevertheless, random splitting and shuffling in the context of traffic classification disrupt the temporal and structural dependencies in real-world network traffic, potentially leading to overly optimistic performance estimates and simplifying the tasks. Therefore, we describe below two primary evaluation protocols that should be considered in traffic classification:

\subsubsection{Disjoint Entities or Environments in Test and Train Sets}
Ensuring that the training and testing datasets do not contain data from overlapping network entities or entire environments is a viable approach in the experimental protocol design. It supports the practical goal of a machine learning algorithm to generalize across previously unseen traffic of a similar type and category. Unfortunately, a disjoint data split requires appropriate support from the dataset, ideally in the form of data provenance~\cite{werder2022establishing}. 

The UCDAVIS19 dataset serves as one example facilitating a disjoint split: while the training set and the Script testing subset are generated similarly, the Human testing subset is created through human interaction with behavioral variability and includes no redundancy with the training set (when considering all packet series metadata).

Some existing datasets, such as the widely used CIC-DoHBrw, can also be redefined to support this approach. Instead of applying random shuffling and splitting for model training and evaluation---as originally proposed by MontazeriShatoori et al.\cite{dataset_CICDoHBrw2020} and followed by subsequent studies---the dataset can be partitioned using disjoint IP addresses for both positive and negative classes. For instance, communication with some DoH resolvers can be included in the training set, while the model is evaluated on unseen communications with different resolvers. This reformulated task definition not only reflects practical deployment more accurately but also introduces greater potential for performance improvement, as demonstrated in \tabref{tab:dohbrw-disjoint}.

\begin{table}[ht]
    \centering
    \caption{Original random shuffle compared to disjoint split on DoHBrw2020 dataset.}
    \begin{tabular}{lccr}
        \toprule
        \textbf{Task} & \textbf{Best Input Space} & \textbf{Max Acc.} & \textbf{$\Delta$} \\ 
        \midrule
        Random shuffle & 98.66 & 98.79 & 0.13 \\  
        Disjoint split & 79.61 & 98.59 & 18.98 \\ 
        \bottomrule
    \end{tabular}
    \label{tab:dohbrw-disjoint}
\end{table}

Such an approach may encourage researchers to propose more generalizable methods, such as that of Jerabek et al.~\cite{jerabek2023dns}, which maintained consistently high performance where others failed, as evidenced in a comparative study~\cite{jerabek2024comparative} evaluated across different datasets, including scenarios with environment changes and disjoint splits.

\subsubsection{Time-based Splitting}
\label{sec:time-based-split}
Time-based data splitting models a realistic deployment scenario by separating training and evaluation datasets chronologically, reflecting how data naturally arrives in an operational environment. The experimental protocol maintains temporal consistency, ensuring that the training data always precedes the testing data. This temporal awareness introduces additional challenges, as it accounts for data drift—the phenomenon where data distributions evolve over time due to factors such as updates, the emergence of new services, or other dynamic changes in the environment. 

Nevertheless, time-based splits require datasets organized chronologically, which are relatively scarce. The examples might be CESNET datasets, such as CESNET-TLS22 and CESNET-QUIC22. We evaluated our baseline on the CESNET-QUIC22 dataset, which exhibits a significant data drift during Week 45. As reported by Luxemburk et al.~\cite{Luxemburk2023QUIC}, Google changed the certificates for most of its services in that week, causing shifts in feature distributions and a steep decline in model performance. 

The results of our evaluation are shown in~\tabref{tab:time-based-split}. We opted for the 1 week evaluation window and Week 44 was used for training, and following weeks 45, 46, and 47 are used as separate testing sets to show model performance degradation over time. As can be seen, the performance is gradually decreasing in the following weeks. When we compare the results of time-based splits with random splitting, we can see that the random split artificially increases the performance by at least 3.5\%.

\begin{table}[ht]
    \centering
        \caption{Comparison of reported performance between Time-based split and Random and shuffled split. The performance is measured in accuracy. For the time-based split, the whole Week 44 was used for training.}
    \begin{tabular}{c | r l  c}
        \toprule
        \multicolumn{1}{c|}{\textbf{Random Split}} & \multicolumn{2}{c}{\textbf{Time-based Split}} & \multirow{2}{*}{\textbf{$\Delta$}} \\
        Best Input Space & Test Week & Best Input Space   \\
        \midrule
        \multirow{3}{*}{79.41} & Week 45  & 75.53 & $-$3.88 \\
                               & Week 46  & 71.88 & $-$7.53 \\
                               & Week 47  & 70.99 & $-$8.42 \\
        \bottomrule
    \end{tabular}
    \label{tab:time-based-split}
\end{table}

\subsection{Minimizing the Collisions Between the Data Samples From Different Classes}

As illustrated in~\figref{fig:ecdf_packet_sequence}, the proportion of sample collision is naturally higher in communications with fewer packets. Unfortunately, such sparse traffic patterns are characteristic of many malicious activities, which is the primary motivator of traffic classification. Examples include reconnaissance port scans, botnet communications, command-and-control channels, and stealthy data exfiltration. These activities are often deliberately short-lived and crafted to blend in with benign background traffic. As a result, reliably identifying them may require more advanced analytical techniques. We therefore encourage researchers to investigate novel approaches to address the limitations revealed in our study.

For instance, when single-flow classification reaches its limits due to overlaps with benign traffic classes, analyzing sequences of consecutive flows, grouped by protocol or network addresses, could offer a more effective alternative. This approach has the potential to improve detection performance by leveraging temporal or contextual correlations. However, pursuing such approaches requires the creation of new datasets tailored for this purpose. One example is the recently introduced dataset~\cite{koumar2025cesnet}, which captures volumetric data from real-world network traffic.

Moreover, machine learning can be supported by different approaches from traditional network traffic processing, such as signature detection. Such heterogeneous solutions may prove more robust and suitable for real-world deployment, as demonstrated by the hybrid DoH detection method of Jerabek et al.~\cite{jerabek2023dns}, or data-fusion for cryptomining detection proposed by Plny et al.~\cite{plny2022decrypto}. In contrast, relying exclusively on machine learning, especially without a critical understanding of its limitations as highlighted in this work, may lead to suboptimal or even unusable results.

\section{Conclusion}\label{sec:conclusion}

Machine learning has become an essential research area in network traffic classification and threat detection, offering solutions to problems that are otherwise difficult to address. Both traditional models and modern deep learning approaches have demonstrated high performance in this domain. However, recent work by Luxemburk et al.~\cite{Luxemburk2025Universal} challenges this trend, showing that a much simpler method based on k-NN can achieve comparable or even superior results to SOTA techniques. In this study, we confirmed their findings and extended the evaluation to 12 influential datasets commonly used in traffic classification. According to our results, the input-space baseline method performed on average with $-2.88\%$ difference compared to the best SOTA, and outperformed the SOTA in two cases.  

Our analysis of the datasets revealed that the strong performance of the baseline is largely driven by extreme data redundancy in TC datasets---a phenomenon that has, until now, been largely overlooked by the community. Commonly used data splitting methodologies often result in identical or nearly identical samples appearing in both the training and test sets. This overlap leads to an overestimation of performance compared to real-world deployment. Moreover, we observed that the exact same samples are frequently assigned different target class labels, making perfect classification inherently unachievable. On four datasets, the performance of the input-space baseline falls within 1\% of the maximum achievable accuracy, raising concerns about the relevance of these commonly used datasets as well as the validity of methods evaluated on them.

We highlight the necessity of developing evaluation protocols tailored specifically to the TC research domain. A straightforward adoption of methodologies from other machine learning fields proves inadequate, as demonstrated by the findings in this paper. Researchers should prioritize time-based and disjoint splitting strategies or develop new protocols to more accurately reflect real-world deployment conditions. Furthermore, a comparison with the simple input-space baseline should be an integral part of any evaluation pipeline, ensuring that the benefits of novel methods are substantiated. Unfortunately, the evidence presented in this paper raises serious concerns that TC research may be caught in a cycle of illusory progress. Only rigorous evaluation and revision of the used dataset can break the circle and provide genuine advancements applicable to real-world deployment.

\ifCLASSOPTIONcompsoc
  \section*{Acknowledgments}
\else
  \section*{Acknowledgment}
\fi
This research was supported by the Ministry of Education, Youth and Sports of the Czech Republic in the project “e-Infrastructure CZ” (LM2023054). It was also supported by the Grant Agency of the CTU in Prague, grant No. SGS23/207/OHK3/3T/18.

\bibliographystyle{IEEEtran}
\bibliography{main}


\newpage

\section*{Appendix A - Heatmap}

~\figref{fig:ecdf_packet_sequence} provides a heatmap offering a detailed view of duplicate distributions across different flow lengths. In some datasets, such as ISCXVPN2016, UNSW-NB15, and CIC-DoHBrw, the fraction of redundant samples remains relatively stable regardless of sequence length. Mixed-class redundancy, however, exhibits a clearer trend: it declines significantly as flow length increases, although it never fully disappears. This indicates that sample redundancy is not confined to short flows but persists even in longer sequences, including those with 30 or more packets.

\section*{Appendix B - Details for SOTA comparison}

For the sake of completeness, we indicate the exact tables of the referenced manuscripts from which the information on the best-performing classifiers for each dataset was obtained. We refer accuracy measure as $\mathbb{A}$, and weighted $F_1$ score as $\mathbb{F}_1$.

\begin{description}[leftmargin=14pt]
    \item[CESNET-TLS22, $\mathbb{A}$] -- Table 5 of Fauvel et al.~\cite{fauvel2023lightweight}. The best result of 97.2\% is achieved with the LEXNet architecture.
    \item[CESNET-QUIC22, $\mathbb{A}$] -- Table 1 of Luxemburk et al.~\cite{Luxemburk2023QUIC}. The best result of 80.87\% is achieved with LightGBM.
    \item[ISCXVPN2016, $\mathbb{A}$] -- Table 3 of Nascita et al.~\cite{Nascita2023Embeddings}. The best results are achieved with the DISTILLER-Embeddings architecture, which uses payload as model input. For a payload-less model comparison, we use 1D-CNN (PSQ) of the same table.
    \item[MIRAGE19, $\mathbb{F}_1$] -- Table 7 of Wang et al.~\cite{wang2024data}. Deltas from this table need to be added to the baseline performance of 75.43\%. The best result is ``MaskedStack ($p = 0.7$)'' with 80.06\% (75.43\% + 4.63\%).
    \item[MIRAGE22, $\mathbb{F}_1$] -- Table 7 of Wang et al.~\cite{wang2024data}. Deltas from this table need to be added to the baseline performance of 94.92\%. The best result is ``MaskedStack ($p = 0.3$)'' with 97.18\% (94.92\% + 2.26\%).
    \item[UTMOBILENET21, $\mathbb{F}_1$] -- Table 8 of Finamore et al.~\cite{finamore2023replication}. The best result for the \textgreater10pkts version is ``Time shift'' with 81.91\%.
    \item[UCDAVIS19, $\mathbb{A}$] -- Table 7 of Finamore et al.~\cite{finamore2023replication} containing results for an enlarged training set. The best result for \textit{human} is ``SimCLR + fine-tuning'' with 80.45\%, and for \textit{script}, it is ``Packet loss'' with 98.63\%.
    \item[AppClassNet, $\mathbb{A}$] -- Table 3 of Wang et al.~\cite{dataset_AppClassNet}, which is the paper introducing the dataset. The best accuracy of 88.3\% on the public version of AppClassNet is achieved with random forest.
    \item[UNSW-NB15, $\mathbb{A}$] -- Table A.9 of Koumar et al.~\cite{Koumar2023NETISA}. The paper introduces a new feature vector based on time-series features. Authors then used this feature vector for the LightGBM model, which yielded reported accuracy scores.
    \item[EdgeIIoTset, CIC-IDS-2017, $\mathbb{A}$] -- Table 3 at Koumar et al.~\cite{koumar2023network}; authors use time-series based features with XGBoost model.
    \item[CIC-DoHBrw, $\mathbb{A}$] -- Table 4 of Mitsuhashi et al.~\cite{mitsuhashi2021identifying}. Accuracy of filtering DoH (binary task), with XGBoost model, using rather classical statistical extension of flow.
\end{description}

\end{document}